\documentclass{article}

\PassOptionsToPackage{numbers, compress}{natbib}

\usepackage{hyperref}

\usepackage[preprint]{neurips_2023}

\usepackage[utf8]{inputenc} 
\usepackage[T1]{fontenc}    
\usepackage{hyperref}       
\usepackage{url}            
\usepackage{booktabs}       
\usepackage{amsfonts}       
\usepackage{nicefrac}       
\usepackage{microtype}      
\usepackage{xcolor}         
\usepackage{multirow}
\usepackage{makecell}

\usepackage{tikz}
\usetikzlibrary{shapes.multipart, arrows.meta, positioning, fit}
\usepackage{enumitem}
\usepackage{subcaption}
\usepackage{amsmath}
\usepackage{pgfplots}

\newcommand{\sysname}{\texttt{SkipDecode}}

\newcommand{\squishlist}{
   \begin{list}{$\bullet$}
    { \setlength{\itemsep}{0pt}      \setlength{\parsep}{3pt}
      \setlength{\topsep}{3pt}       \setlength{\partopsep}{0pt}
      \setlength{\leftmargin}{1.5em} \setlength{\labelwidth}{1em}
      \setlength{\labelsep}{0.5em} } }

\newcommand{\squishlisttwo}{
   \begin{list}{$\bullet$}
    { \setlength{\itemsep}{0pt}    \setlength{\parsep}{0pt}
      \setlength{\topsep}{0pt}     \setlength{\partopsep}{0pt}
      \setlength{\leftmargin}{0.5em} \setlength{\labelwidth}{1em}
      \setlength{\labelsep}{0.2em} } }

\newcommand{\squishend}{
    \end{list}  }

\title{SkipDecode: Autoregressive Skip Decoding with Batching and Caching for Efficient LLM Inference}

\author{
    \textbf{Luciano Del Corro }\hspace{1cm} 
    \textbf{Allie Del Giorno}\hspace{1cm} 
    \textbf{Sahaj Agarwal}\hspace{0.8cm}\\\\
    \textbf{Bin Yu}\hspace{0.8cm}
    \textbf{Ahmed Awadallah}\hspace{0.8cm}
    \textbf{Subhabrata Mukherjee} \\\\
    Microsoft Research\\
    \texttt{\{ldelcorro,adelgiorno,sahaj.agarwal\}@microsoft.com},\\
    \texttt{\{v-ybi,ahmed.awadallah,subhabrata.mukherjee\}@microsoft.com}
}

\begin{document}

\maketitle

\begin{abstract}
Autoregressive large language models (LLMs) have made remarkable progress in various natural language generation tasks. However, they incur high computation cost and latency resulting from the autoregressive token-by-token generation. To address this issue, several approaches have been proposed to reduce computational cost using early-exit strategies. These strategies enable faster text generation using reduced computation without applying the full computation graph to each token. While existing token-level early exit methods show promising results for online inference, {\em they cannot be readily applied for batch inferencing and Key-Value caching}. This is because they have to wait until the last token in a batch exits before they can stop computing. This severely limits the practical application of such techniques. 
In this paper, we propose a simple and effective token-level early exit method, {\sysname}, designed to work seamlessly with batch inferencing and KV caching. It overcomes prior constraints by setting up a singular exit point for every token in a batch at each sequence position. It also guarantees a monotonic decrease in exit points, thereby eliminating the need to recompute KV Caches for preceding tokens. Rather than terminating computation prematurely as in prior works, our approach bypasses lower to middle layers, devoting most of the computational resources to upper layers, allowing later tokens to  benefit from  the compute expenditure by earlier tokens.
Our experimental results show that {\sysname} can obtain 2x to 5x inference speedups with negligible regression across a variety of tasks. This is achieved using OPT models of 1.3 billion and 6.7 billion parameters, all the while being directly compatible with batching and KV caching optimization techniques.

\end{abstract}

\section{Introduction}

Autoregressive large language models (LLMs), such as the GPT~\cite{Radford2018ImprovingLU} and OPT~\cite{Zhang2022OPTOP} family, have demonstrated strong performance across a wide range of tasks~\cite{ Radford2019LanguageMA, Raffel:2020, NEURIPS2020_1457c0d6}. However, thee also have high computational cost and latency requirements resulting from token-by-token generation. Token-level early exit~\cite{schuster2022confident, SunLZGWHNXHQ22} has emerged as a promising technique to alleviate these limitations by allowing tokens to cease computation as soon as their hidden states reach saturation~\cite{schuster2022confident}.

Although current methodologies exhibit theoretical advantages, their practical implementation is somewhat restricted since they are not compatible with batch inferencing and KV caching techniques, which are widely used to speed up inference in practice. This is mainly due to the necessity of prolonging computation until the final token in a batch for each position is thoroughly processed. This effectively limits improvements to the exit position of the most computation-heavy token. Additionally, token-level exit strategies, which depend on dynamic indicators like learned classifiers for defining exit points, don't provide any assurance concerning computational expenses, such as the worst-case cost relative to the computation performed by the full-sized network. A further practical difficulty arises in the form of Key-Value (KV) caching of prior tokens, which requires updating if the current token exits later than the others.

\begin{figure}[t]
    \centering
    \begin{subfigure}{.45\textwidth}
        \centering
        \includegraphics[trim=0cm 0 0cm 2cm,clip,width=\linewidth]{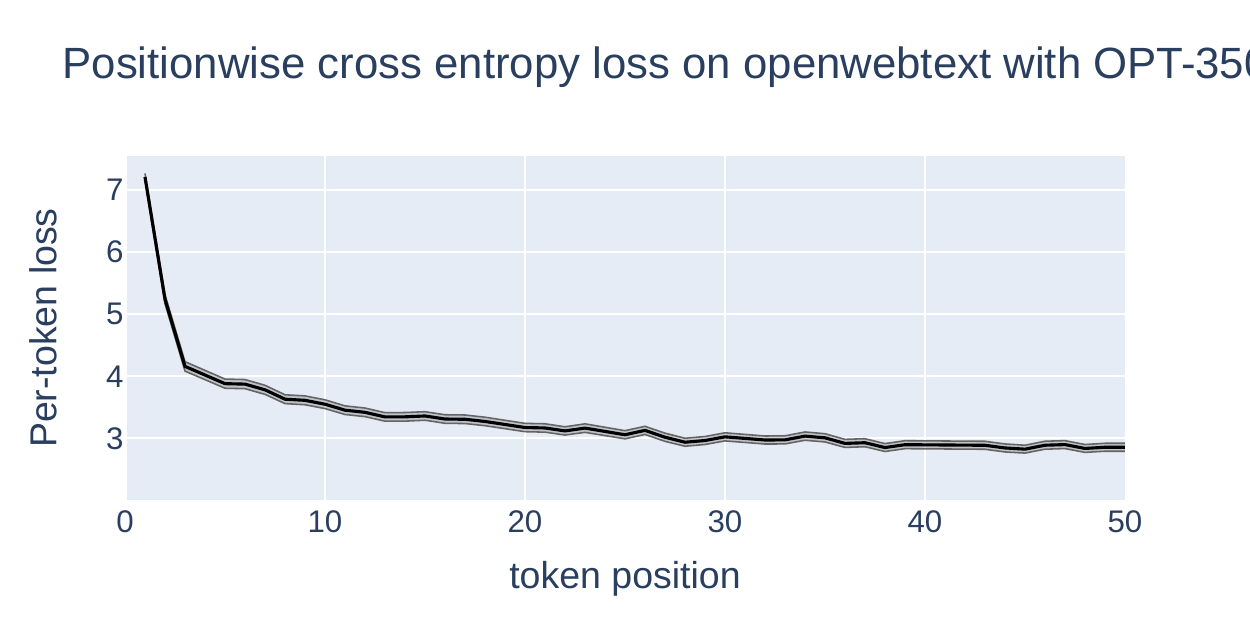}
        \caption{\textbf{OPT-350m on OpenWebText}. Average loss per token position shows a strong monotonically-decreasing trend for general text.  
        }        \label{fig:monotonic_xent_openwebtext}
    \end{subfigure}%
    ~
    \begin{subfigure}{.45\textwidth}
        \centering
        \includegraphics[trim=0cm 0 0cm 2cm,clip,width=\linewidth]{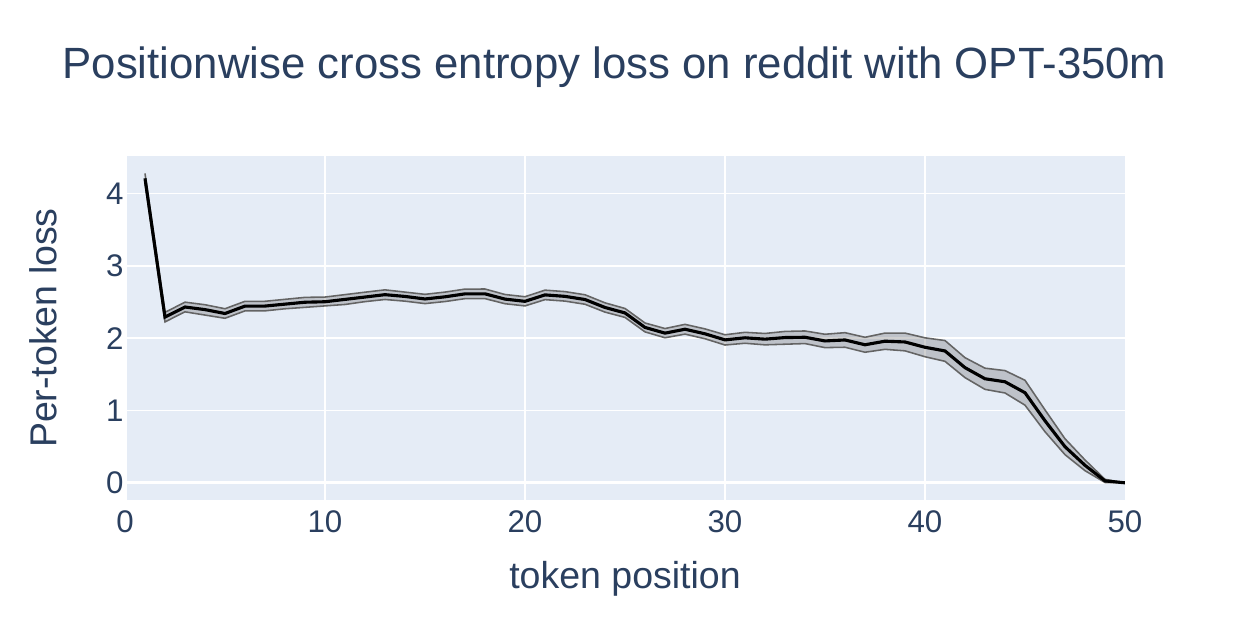}
        \caption{\textbf{OPT-350m (finetuned) on Reddit-TLDR}. Average loss per token position. Decreasing trend but with a different function.}
        \label{fig:monotonic_xent_reddit}
    \end{subfigure}
    \vspace{-0.5em}
    \caption{Average loss per token position (black) during the forward pass of OPT-350m model on a general and a task-specific dataset. Grey represents the 95\% confidence interval on the mean.}
    \label{fig:oracle}
    \vspace{-1.5em}
\end{figure}

In this paper, we present a novel token-level early exit method, called {\sysname}, which overcomes these limitations, while maintaining a controlled computational budget. Our approach establishes a unified exit point for all tokens within a batch at a specific sequence position. We further capitalize on the observation that words towards the end of a sequence are generally easier to predict due to more contextual information. This allows us to design our generation policy with monotonically decreasing exit points as the sequence progresses, with the assumption that subsequent tokens demand less computational effort. Figure \ref{fig:oracle} shows the decreasing loss per token position: Predictions at the beginning of the sequence register higher entropy in contrast to the tokens that appear later. This motivates the use of increased computational effort upfront to minimize the mistakes, and therefore prevent the cascading of errors; whereas we can reduce the computational processing later as the sequences become more predictive. Our strategy with monotonically decreasing exit points also eliminates the necessity to recalculate Key-Value (KV) caches for preceding tokens, substantially reducing computational cost.

Tokens exiting at distinct layers are unable to benefit from all the information generated by previous tokens that exit at later positions, leading to wasted computation and loss of contextual information. To address this issue, the early exit mechanism in {\sysname} leverages the entire computation performed by all tokens, resulting in a substantial improvement in the speedup-task performance trade-off. Rather than abruptly ending computation, our approach bypasses lower layers and primarily allocates the computational budget to upper layers, enabling rightward tokens to benefit from the computational resources employed by leftward tokens effectively.

\begin{figure}[t!]
  \centering
  \begin{subfigure}[t]{0.49\textwidth}
    \tiny
    Practical Blockers (Existing):
    \begin{itemize}[leftmargin=*]
      \item {\bf Batching:} Computational cost defined by the last exit token
      \item {\bf KV Caching:} If next token exits later than previous one, we need to recompute KV values for previous tokens.
      \item {\bf Computational Efficiency:} If next token exits earlier, it does not attend full computation of previous token.
      \item {\bf Cost Uncertainty:} Worst case scenario (for a bad token exit, e.g., from last layer) equivalent to processing the whole network.
    \end{itemize}
    \includegraphics[width=\textwidth]{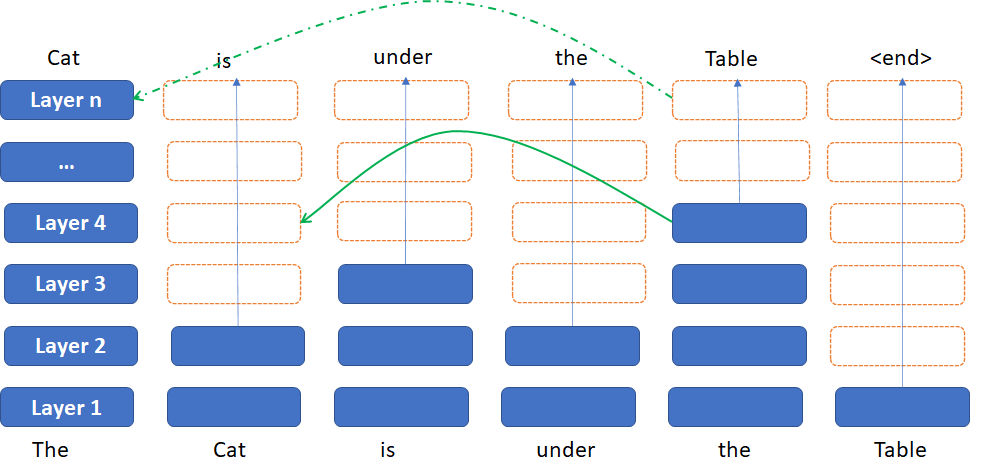}
    \vspace{0.008cm}
    \caption{Early Termination}
  \end{subfigure}
  \hfill
  \begin{subfigure}[t]{0.49\textwidth}
    \tiny
    Solutions (Ours):
    \begin{itemize}[leftmargin=*]
      \item {\bf Batching:} Exit per position per batch (column-wise).
      \item {\bf KV Caching:} Next column has to exit earlier than previous column. Leftwards tokens are more difficult to generate.
      \item {\bf Computational Efficiency:} Spend most of computational budget on top layers. Implicitly attends the full computation of previous tokens.
      \item {\bf Cost Uncertainty:} Static policy (no surprises), computation costs are predefined.
    \end{itemize}
    \includegraphics[width=\textwidth]{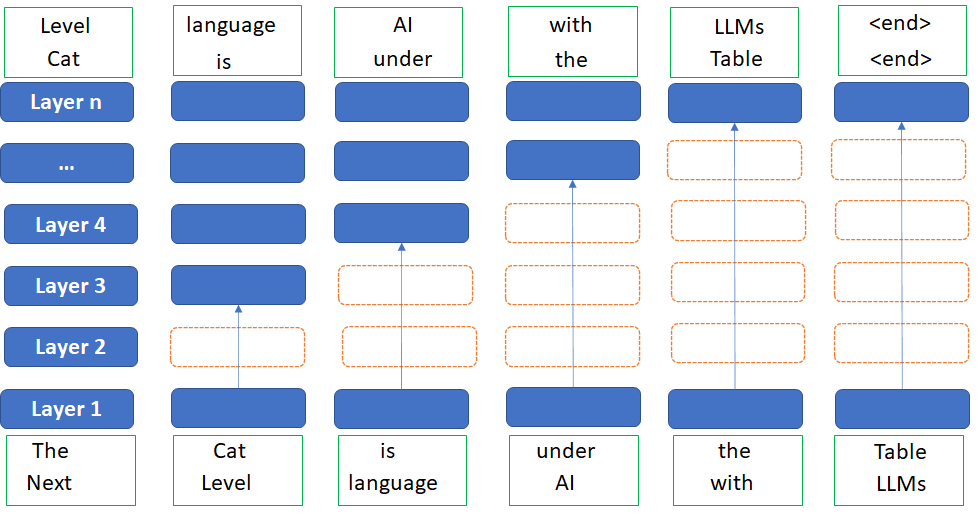}
    \caption{Skipping}
  \end{subfigure}
  \caption{Overcoming practical blockers in token level early exit for language generation.}
  \label{skipping_early}
  \vspace{-1.5em}
\end{figure}

Our technique  {\sysname} (overview in Figure \ref{skipping_early} and Table \ref{tab:comparison}) is able to avoid performance degradation up to the hidden state saturation point. We experiment with up to 5x speedup over 1.3 billion, and 6.7 billion OPT Transformer models on three benchmark generation datasets. We also solve practical problems like batching and KV caching while  maintaining a controlled and predictable computational budget. Our method makes it easier to use LLMs on devices with limited resources and helps to democratize AI.

\begin{table}[h]
\centering
\footnotesize
\begin{tabular}{cccccccc}
\toprule
\textbf{Method} & \textbf{Model} & \textbf{Generation} & \textbf{Token} & \textbf{Batching} & \textbf{KV-} & \textbf{Full} & \textbf{Controlled} \\
& \textbf{Type} & & \textbf{Level} & & \textbf{Caching} & \textbf{Attention} & \textbf{Comp. Cost} \\
\midrule
CALM & Enc-Dec & \checkmark & \checkmark & $\times$ & $\times$ & $\times$ & $\times$ \\
{\sysname} & Dec Only & \checkmark & \checkmark & \checkmark & \checkmark & \checkmark & \checkmark \\
\bottomrule
\end{tabular}
\caption{Comparison of CALM and {\sysname}. {\sysname} supports batching and KV caching for increasing inference efficiency with controlled computational budget.}
\label{tab:comparison}
\vspace{-2em}
\end{table}

\section{Addressing Practical Challenges in Token-Level Early Exit Strategies}

In this section, we discuss the practical challenges associated with token-level early exit methods in LLMs and present {\sysname} to address these limitations.

\subsection{Optimizing Batched Inference with Unified Exit Points}

Batched inference is widely used to enhance computational efficiency by simultaneously processing multiple input samples. This approach capitalizes on the parallelism offered by hardware, such as GPUs and TPUs, to reduce latency and improve throughput.

However, when applying existing token-level early exit strategies~\cite{schuster2022confident} to batched inference, challenges arise primarily due to the varying exit points of tokens within a batch. Given that tokens exit at diverse layers, it's necessary to persist computation until the final token of each batch member and each position is processed. This diminishes the benefits that would otherwise be realized when the batch size exceeds one, thus undermining the potential advantages of parallel computation.

To tackle this, we suggest a method that designates a fixed positionwise exit point for every token in a batch at a given sequence position. This strategy ensures that the processing of all batched tokens at a certain sequence position concludes at the same time. As a result, it assures that all theoretical benefits observed during evaluation are fully actualized during generation for non-trivial batching scenarios.

Let $B$ be the batch size, and $N$ the sequence length. We construct the batches column-wise using tokens at a specific position across all the instances. Considering $t_{s,n}$ to be the token in sequence $s$ at position $n$, a given batch consists of tokens from $t_{(\cdot),n}$. 
Let $L(t_{b,n})$ be the layer at which token $t_{b,n}$ exits. We ensure that $\forall n \in [1, N], \forall b_1, b_2 \in [1, B], L(t_{b_1,n}) = L(t_{b_2,n})$. Further, the autoregressive decoding for token generation ensures that the columns are processed left to right such that the computation for tokens at position $n$ can utilize all the network computation from processing those at the previous position $n-1$.

\subsection{KV Caching and Monotonically Decreasing Exit Points}

Key-Value(KV) KV caching is a critical optimization technique for efficiently executing attention mechanisms in Transformer models. By storing the computed keys and values for previously processed tokens, the model can significantly reduce redundant computations when attending to the same context during subsequent steps. This optimization leads to faster inference times.

Yet, when utilizing token-level early exit strategies , the different exit points of tokens in a sequence present another challenge. Specifically, there's a requirement to recompute the Key-Value (KV) caches for preceding tokens if the current token exits at a higher layer. This necessary recalculation escalates computational workload and undermines the advantages of early exit methods, as the computation of each preceding token is bounded by the computation of later tokens.

Our proposed solution, which assigns a unified exit point to all tokens in a batch at a given position, effectively addresses this challenge. By ensuring that batched exit points are monotonically decreasing as the sequence progresses, we guarantee that previous tokens have performed at least as much computation as the current one, thereby trivially avoiding the need for any extra computation. The right plot in figure \ref{skipping_early} shows how every layer can attend to leftward attention layers without any re-computation or change in the architecture. 

The underlying rationale is that next-word prediction at the beginning of a sequence is more challenging due to limited context, and therefore earlier tokens will benefit from later exit points in the computation graph. Prior work~\cite{schuster2022confident} have already showed that noise or perturbations have a greater impact on the overall task performance when the perturbation happens in the earlier tokens resulting in cascading of errors due to autoregressive generation. Moreover, as the context grows with the sequence, the later tokens become more predictive with more context resulting in their hidden states saturating faster (i.e. hidden states have limited variance across layers). Therefore, later tokens require less computation and thus enabling a more efficient use of computational resources~\cite{Holtzman2020The}. We demonstrate this intuition in Figure \ref{fig:oracle}, where earlier tokens in a sequence have higher losses and are more difficult to generate in contrast to the ones appearing later that are more predictive.

\subsection{Controlling Computational Budget with Batched Exit Points}\label{computational_budget}

Traditional early exit techniques typically learn exit points for individual tokens \cite{schuster2022confident}. However, apart from the limitations mentioned in the previous subsections, controlling the computational budget can be challenging. Usually, a classifier is used to decide whether a token should exit at a specific layer, resulting in the worst-case computational budget scenario being close to the cost of using the full network (for instance, bad exit point close to the last layer).

We address this issue by pre-specifying maximum and minimum exit points (the maximum and minimum number of layer that each token should go through), which controls the computational cost via the number of active model parameters. Exit points across the sequence are assigned in such a way that no token exceeds the maximum nor falls below the minimum exit point keeping the total computational cost bounded. Additionally, as explained earlier, the assignment of exit points across the sequence is required to be monotonically decreasing. This implies that the first token will be assigned the maximum exit point, and the last token, according to the maximum length parameter, will be assigned the minimum exit point.

A predefined function progressively designates exit points to the tokens in the sequence. This function can adopt multiple forms and serves as an additional hyperparameter of the model, managing the computational expenditure. In the evaluation phase, we conduct tests using linear decay bounded by a a minimum and a maximum number of layers. Note that employing other functions (such as power-law) could lead to more significant accelerations and will be the subject of future studies.

Formally, consider a sequence and network with hyper-parameters: $\text{sequence\_length}$, $\text{min\_exit\_layer}$, $\text{max\_exit\_layer}$, $\text{num\_decoder\_layers}$, and $\text{prompt\_size}$. We define an array $\text{token\_idx}$ as:

\[
\text{token\_idx}[i] = 
\begin{cases} 
\text{num\_decoder\_layers} & \text{if } i < \text{prompt\_size} \\
(1 - t_i) \times \text{max\_exit\_layer} + t_i \times \text{min\_exit\_layer} & \text{if } i \geq \text{prompt\_size}
\end{cases}
\]

where $t_i = \frac{i - \text{prompt\_size}}{\text{sequence\_length} - \text{prompt\_size}}$.

\begin{figure}[h]
\centering
\begin{tikzpicture}[scale=0.65,
  declare function={
    func(\x)= (\x < 20) * 12 +
              and(\x >= 20, \x <= 100) * ((1 - (\x - 20) / (100 - 20)) * 8 + ((\x - 20) / (100 - 20)) * 2);
  }
]
\begin{axis}[
    title={Exit Layer vs Sequence Position},
    xlabel={Sequence Position},
    ylabel={Exit Layer},
    xmin=0, xmax=100,
    ymin=0, ymax=12,
    xtick={0,20,100},
    xticklabels={0,prompt\_len,max\_len},
    ytick={2,8,12},
    yticklabels={min\_layer,max\_layer,num\_decoder\_layers},
    legend pos=north east,
    ymajorgrids=true,
    grid style=dashed,
]

\addplot[
    color=red,
    ultra thick, 
    domain=0:100,
    samples=100,
]{func(x)};

\end{axis}
\node at (2.5,2) {\small Computational Budget};
\end{tikzpicture}
\vspace{-0.5em}
\caption{Linear decay of exit layer with respect to the sequence position.}
\label{fig:linear_decay}
\vspace{-0.5em}
\end{figure}
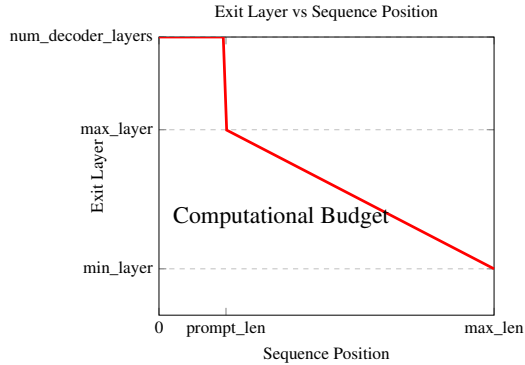

In the above design, we can process all the tokens in the prompt with the full computational power of the network (i.e. using all the decoder layers). This can be done efficiently with batching since there is no generation involved in prompt processing. As soon as we start autoregressive token generation from $\text{prompt\_len}+1$, we start decaying the number of active layers acting on any token bounded by the max and min layers pre-specified by the computational budget.

\subsection{Skipping vs. Early Termination}

Early termination-based methods allow tokens to exit the model early once their hidden states have saturated\cite{schuster2022confident}. However, token-level early termination can present problems in generative models when previous tokens exit earlier than the later tokens. In this scenario, the later tokens are unable to benefit from the extra computation performed by the former tokens via the attention mechanism, effectively under utilizing the available context.

To overcome this limitation, we propose performing skipping instead of early termination. We ensure that the computational budget for each token is allocated to higher layers of the model. Now regardless of the exit points, tokens will be able to attend to the top layers of all previous tokens effectively attending to all the available context. 

To bridge the representation gap between the initial embedding layer and the top layers, we introduce the concept of warmup layers. Warmup layers represent the initial computation that will be performed on the $x$ bottom layers before skipping to the top $y$ layers to exhaust the rest of the computational budget. In experiments we observe that this approach effectively reduces the distance between the token embeddings and the top layer's hidden states. In experimental evaluation, we consistently found the number of warmup layers to be 1 that worked the best across all settings.

\section{Evaluation}

We demonstrate our techniques with OPT~\cite{Zhang2022OPTOP} decoder-only language models of 1.3b and 6.7b parameters (24 and 32 layers respectively) on three benchmark text generation datasets: E2E~\cite{novikova2017e2e}, Reddit-TLDR~\cite{volske2017tl}, and CNN-DM~\cite{hermann2015teaching}. We implement {\sysname} using the metaseq codebase\footnote{https://github.com/facebookresearch/metaseq}. 

\subsection{Experimental Design}

Given a pre-trained LLM (base model) assigned a $1\times$ speedup, our goal is to reduce the amount of computation performed by the network during inference using {\sysname}. We evaluate our method with configurations corresponding to $2\times, 3\times, 4\times$ {and} $5\times$ speedups. {\bf Note that speedups are reported relative to the base model that intrinsically supports batching and KV caching.} This speedup comparison is different from prior early-exit works that consider a weaker base model with a batch size of $1$ and no KV caching.

Different configurations of the maximum and minimum layer per token, as outlined in Section~\ref{computational_budget}, can achieve different amounts of speedup. We determine the optimal combination of maximum and minimum layers, along with the warm-up layers and learning rate, for each specified speedup through hyper-parameter tuning on the e2e dataset. We select the optimal one based on the perplexity metric on the {\bf validation set}. It should be noted that the actual speedup may slightly vary during generation as it's impossible to predict in advance the number of tokens that the model will generate. However, the computational budget is strictly bounded by the minimum layer assigned to the maximum sequence length for each dataset. The configurations used are presented in Table \ref{table:congifurations}.

\begin{table}[]
\footnotesize
\centering
\begin{tabular}{ccccccc}
\toprule
\textbf{Original / Base} & \textbf{Target} & \textbf{\#Target Avg} & \textbf{\#Warm up} & \textbf{\#Min} & \textbf{\#Max} \\
\textbf{Number of Layers} & \textbf{Speedup ($\times$)} & \textbf{Layer} & \textbf{Layer} & \textbf{Layer} & \textbf{Layer} \\
\midrule
\multirow{4}{*}{32 (6.7B)} & 2 & 16 & 1 & 11 & 22 \\
 & 3 & 11 & 1 & 8 & 14 \\
 & 4 & 8 & 1 & 6 & 10 \\
 & 5 & 6.5 & 1 & 5 & 8 \\
\midrule
\multirow{4}{*}{24 (1.3B)} & 2 & 12 & 1 & 8 & 16 \\
 & 3 & 8 & 1 & 6 & 10 \\
 & 4 & 6 & 1 & 5 & 7 \\
 & 5 & 5 & 1 & 4 & 6 \\
\bottomrule
\end{tabular}
\caption{{\sysname} configurations for different target speed-ups w.r.t Base OPT (1.3B and 6.7B) obtained using the E2E validation set corresponding to the least perplexity.}
\label{table:congifurations}
\vspace{-2em}
\end{table}

For training, we used the median training prompt length from each dataset for all instances, ensuring that all layers are processed to mimic the desired generation behavior as illustrated in figure \ref{fig:linear_decay}.

It's worth noting that our approach is effective yet simple and easy to implement. Besides the token skipping policy, it does not necessitate any additional modifications to the transformer architecture, either during training or generation.

\subsection{Datasets}

We perform experiments on three benchmark datasets. Examples of generation on each dataset are shown in Table \ref{table:dataset_examples}. For generation, in all cases we employ a beam of 1, top-sampling of 0.7, and a temperature of 0.3. For training we sweep over learning rates in the range of 2e-4 to 8e-6.

\begin{table}[hb]
\centering
\tiny
\begin{tabular}{l p{6.5cm} p{2.5cm} p{2cm}}
\toprule
\textbf{Dataset} & \textbf{Context} & \textbf{Response 2x} & \textbf{Response 5x} \\
\midrule
E2E & name[Blue Spice], eatType[coffee shop], customer rating[average], near[Burger King] & The Blue Spice coffee shop located near Burger King has been rated average by customers. & Blue Spice is a coffee shop near Burger King. It has an average customer rating and is located near the Burger King. \\
\midrule
Reddit-TLDR & "SUBREDDIT: r/relationships TITLE: This guy I've been casually dating [18M] doesn't want to have a relationship with me [18F] because he's going to college in the fall POST: Here's a bit of context for y'all: We both met freshmen year in our school's theatre program. At the end of freshman year, I transferred to...
& Guy I've been casually dating wants to break up with me because he's going to university in the fall and I have to stay in high school for another year. & Guy I've been dating has been dating for a while, he's going to university in the fall, I'm crushed and don't know how to proceed.\\
\midrule
CNN-DM & (CNN)The terrorist group Al-Shabaab has claimed an attack on Garissa University College in eastern Kenya, in which many people have been killed and still more taken hostage. The attack is another step in the ongoing escalation of the terrorist group's activities, and a clear indicator that the security situation in East Africa is deteriorating fast. Somalia-based Al-Shabaab has been behind a string of recent attacks in Kenya, the most well-known of them being the massacre at the Westgate Shopping Centre in Nairobi in 2013. Cross-border raids into Kenya by the group, however, date back to 2011. Al-Shabaab incursions triggered a military... 
& Al-Shabaab claims attack on Garissa University College in Kenya .Attack is another step in the ongoing escalation of terrorist group's activities .Al-Shabaab has been behind a string of recent attacks in Kenya .The group is predominantly driven by the same radical interpretation of the Koran as al-Qaeda .
& Al-Shabaab has claimed an attack on Garissa University College in Kenya .Al-Shabaab has been behind a string of recent attacks in Kenya .Al-Shabaab has been behind a string of recent attacks in the region. \\
\bottomrule
\end{tabular}
\caption{Snapshot of dataset and model responses.}
\label{table:dataset_examples}
\vspace{-2em}
\end{table}

{\bf E2E~\cite{novikova2017e2e}.} The task is to convert structured information from key-value pairs into fluent text. It is relatively small, comprising of 42061 training samples, 4672 evaluation samples, and 4693 test samples. The median prompt contains 38 tokens. We set a maximum sequence length of 160 and a maximum prompt length of 60, both for training and generation. The effective batch size is 256. We use a breakline to separate the prompt and completion with 650 warm-up steps and 8 epochs.

{\bf Reddit-TLDR~\cite{volske2017tl}.} A summarization dataset that includes a training size of 117,000 samples, an evaluation size of 6450, and a test size of 6550 samples. The median training prompt is 348 tokens long. We utilize 200 warm-up steps, an effective training batch of 32, and 3 epochs. The maximum prompt length is set at 512 tokens, and the maximum sequence length at 1024 tokens. The separator between context and completion is "\textbackslash nTl;dr\textbackslash n".

{\bf CNN Daily Mail~\cite{hermann2015teaching}.} Requires writing a summary given an article as input. It is a large dataset with a training size of 287,113 samples, an evaluation size of 13,368, and a test size of 11,490 samples. The median train prompt length is 788 tokens. We set the maximum sequence length at 2048 and the maximum prompt length at 1024. The warm-up updates are set to 650, the effective batch size is 32, and we train for 2 epochs. The separator between context and completion is "\textbackslash nTl;dr\textbackslash n".

\subsection{Key Results}
\begin{figure}[htbp]
    \centering
    \begin{subfigure}{.33\textwidth}
        \centering
        \includegraphics[width=\linewidth]{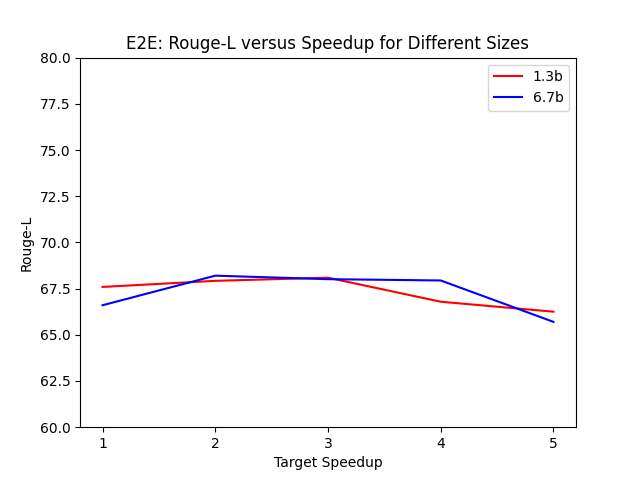}
        \caption{E2E}
        \label{fig:e2e}
    \end{subfigure}%
    \hfill
    \begin{subfigure}{.33\textwidth}
        \centering
        \includegraphics[width=\linewidth]{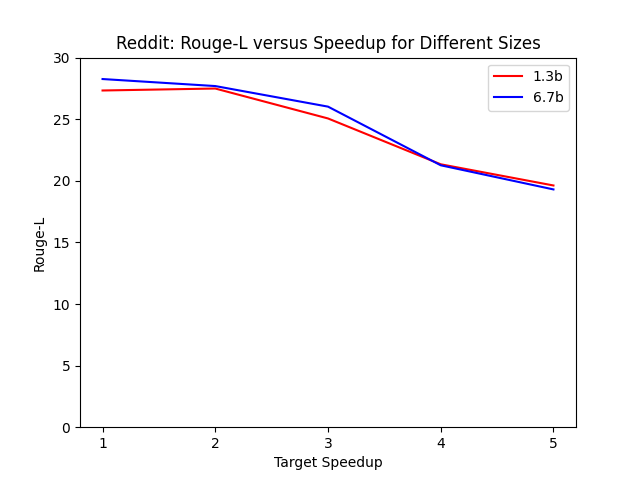}
        \caption{Reddit}
        \label{fig:reddit}
    \end{subfigure}
    \hfill
    \begin{subfigure}{.33\textwidth}
        \centering
        \includegraphics[width=\linewidth]{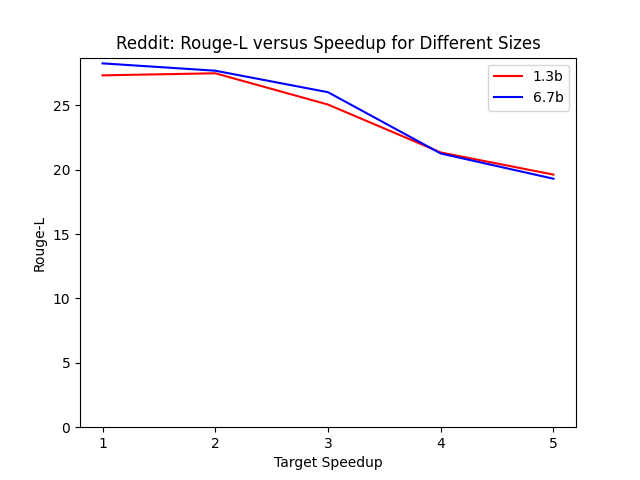}
        \caption{CNN\_DM}
        \label{fig:cnn_dm}
    \end{subfigure}
    \vspace{-0.3em}
    \caption{Rouge-L vs inference speedup for 1.3B and 6.7B OPT models. Speedup is computed over base model ($1\times$) that inherently supports batching and KV caching in contrast to prior work considering a weaker base model without batching as reference.}
    \label{fig:comparison}
    \vspace{-1.4em}
\end{figure}

Results are presented in Table \ref{unified_results}. {\sysname} demonstrates significant improvement in computational efficiency for each dataset and model size. As depicted in Figure \ref{fig:comparison}, there is no noticeable performance degradation from $1\times$ (base model) to $2\times$ speedup, after which there is steady decline in performance with increasing speedups. We hypothesize that this is due to the tokens reaching the hidden state saturation point, beyond which further computation reduction leads to performance degradation. This pattern is consistent across datasets. We notice a delayed degradation in E2E, while CNN-DM starts to degrade more quickly given its relative difficulty.

{\bf E2E}
{\sysname} As the target speedup increases from $1\times$ to $5\times$, the average number of decoder layers active in the generation process reduces, signifying a decrease in the computation load. Interestingly, all the task measures corresponding to the Bleu, Rouge-L, and Bert-F scores remain relatively steady with a minor decline as the target speedup increases. This indicates that our method can accomplish significant speedups with minimal degradation for certain task settings.

{\bf Reddit} Similar to the others, the average generation layer decreases as the target speedup increases. However, the performance metrics such as Bleu, Rouge-L, and Bert-F scores display more significant reductions compared to the E2E dataset, given the relative difficulty of this task. Wile our method still achieves significant speedups, the trade-off in terms of task performance is more noticeable.

{\bf CNN-DM} The results follow a similar trend to the previous: as the target speedup increases, the average generation layer decreases, indicating reduced computational requirements. However, the performance metrics such as Bleu, Rouge-L, and Bert-F scores drop more significantly as the target speedup increases. While our approach can achieve substantial speedups, the trade-off in task performance is more pronounced, as the hidden state saturation is reached earlier.

In conclusion, our method consistently demonstrates an ability to decrease computational demands across all datasets and model sizes, effectively determining the hidden state saturation point. The impact on task performance, as measured by Bleu, Rouge-L, and Bert-F scores, varies depending on the specific dataset. However, in all instances, our method shows a favorable balance between speedup and task performance, reaching a $2\times$ speedup with almost no degradation in all cases. This balance can be effectively exploited as {\em our approach adeptly handles practical challenges like batching and KV caching while maintaining a controlled and predictable computational budget}. 

\begin{table}[htbp]
\centering
\footnotesize
\begin{tabular}{cccccccc}
\toprule
\textbf{Dataset} & \textbf{Size} & \textbf{Target} & \textbf{\#Target Avg} & \textbf{\#Gen Avg} & \textbf{Bleu} & \textbf{Rouge-L} & \textbf{Bert-F} \\
 &  & \textbf{Speedup} & \textbf{Layer} & \textbf{Layer} &  &  &  \\
\midrule
\multirow{5}{*}{E2E} & \multirow{5}{*}{1.3b} & 1 & 24 & 24 & 65.8 & 67.6 & 70.3 \\
 &  & 2 & 12 & 14.7 & 66.3 & 67.9 & 67.8 \\
 &  & 3 & 8 & 9.4 & 66.3 & 68.1 & 67.3 \\
 &  & 4 & 6 & 6.8 & 65.6 & 66.8 & 66.5 \\
 &  & 5 & 5 & 5.8 & 64.2 & 66.3 & 65.2 \\
\cmidrule{2-8}
 & \multirow{5}{*}{6.7b} & 1 & 30 & 30 & 64.2 & 66.6 & 70.8 \\
 &  & 2 & 15 & 20.3 & 65.3 & 68.2 & 67.6 \\
 &  & 3 & 11 & 13 & 65.9 & 68.0 & 67.7 \\
 &  & 4 & 8 & 9.4 & 66.9 & 67.9 & 67.1 \\
 &  & 5 & 6.5 & 7.6 & 64.0 & 65.7 & 65.2 \\
\midrule
\multirow{5}{*}{Redit} & \multirow{5}{*}{1.3b} & 1 & 24 & 24 & 9.0 & 27.3 & 31.9 \\
 &  & 2 & 12 & 15.6 & 8.9 & 27.5 & 32.1 \\
 &  & 3 & 8 & 9.9 & 7.0 & 25.1 & 22.9 \\
 &  & 4 & 6 & 6.4 & 3.9 & 21.3 & 11.5 \\
 &  & 5 & 5 & 5.0 & 3.0 & 19.6 & 7.1 \\
\cmidrule{2-8}
 & \multirow{5}{*}{6.7b} & 1 & 30 & 30 & 9.6 & 28.3 & 33.7 \\
 &  & 2 & 15 & 19.8 & 9.3 & 27.7 & 32.3 \\
 &  & 3 & 11 & 13.7 & 8.0 & 26.0 & 25.3 \\
 &  & 4 & 8 & 9.4 & 5.2 & 21.3 & 9.3 \\
 &  & 5 & 6.5 & 6.5 & 4.0 & 19.3 & 7.4 \\
\midrule
\multirow{5}{*}{CNN-DM} & \multirow{5}{*}{1.3b} & 1 & 24 & 24 & 15.8 & 29.5 & 35.9 \\
 &  & 2 & 12 & 15.6 & 15.0 & 28.9 & 34.8 \\
 &  & 3 & 8 & 8.9 & 7.8 & 23.3 & 20.2 \\
 &  & 4 & 6 & 6.2 & 3.2 & 18.6 & 2.3 \\
 &  & 5 & 5 & 5.3 & 4.0 & 18.1 & 2.5 \\
\cmidrule{2-8}
 & \multirow{5}{*}{6.7b} & 1 & 30 & 30 & 16.3 & 30.2 & 37.1 \\
 &  & 2 & 15 & 21.3 & 15.2 & 29.6 & 35.9 \\
 &  & 3 & 11 & 11.8 & 4.8 & 21.8 & 17.9 \\
 &  & 4 & 8 & 8.5 & 5.4 & 20.2 & 7.9 \\
 &  & 5 & 6.5 & 6.9 & 4.6 & 18.5 & 2.7 \\
\bottomrule
\end{tabular}
\caption{{\sysname} performance on different datasets for varying speedups and base model sizes.}
\label{unified_results}
\vspace{-2em}
\end{table}

\subsection{Comparison to other Methods}
In order to benchmark \sysname, we have adapted two concepts from the CALM framework to function on decoder-only models, though no method is currently available that corresponds directly with \sysname. In both cases we use OPT 1.3b as the base model.

Firstly, we train a multi-layer exit network following the method outlined in \cite{schuster2022confident}, wherein a single model head is trained to exit from each layer. This approach is more akin to an early termination method with truncation as it operates with a fixed policy, running up to a predetermined exit layer applicable to all tokens within the sequence. Notably, this model supports batching and KV Caching.

The second method uses the same model, with an additional application of the CALM's hidden state saturation concept, adapted to function with a decoder-only network (referred as CALM-DEC). However, this network imposes a limitation on the batch size to just one, precluding batching and KV Caching. Consequently, the model must 'back-fill' all KV values for previous tokens as required (in this case, by projecting the last known hidden state at that layer), which adds significant systems overhead. The worst case computational cost of this approach is equivalent to the full network cost.

The adaptive hidden state saturation policy on this network has the standard disadvantages of a non-fixed policy for both batching and computation/time estimates. In addition, its performance degrades strongly with increasing speedups especially on larger decoder-only models for the following reason. 
The KV backfill affects the prompt encoding, which is extremely important for these tasks.  In an encoder-decoder architecture like the T5 model in CALM~\cite{schuster2022confident}, the KV backfill retains the prompt encoding. Whereas decoder-only architectures simultaneously encode and decode their past state, meaning that early termination is more likely to affect the network's understanding of previous context (refer to Appendix). This results in our CALM implementation showing significant degradation for decoder-only models as opposed to the original T5 encoder-decoder implementation.

\begin{table}[h]
\centering
\footnotesize
\begin{tabular}{lcccccccc}
\toprule
 \textbf{Speedup} & & \multicolumn{3}{c}{\textbf{E2E}} & & \multicolumn{2}{c}{\textbf{Reddit-TLDR}} \\
\cmidrule{3-5} \cmidrule{7-8}
 & & \textbf{\sysname} & \textbf{Multi-layer} & \textbf{CALM-DEC} & & \textbf{\sysname} & \textbf{Multi-layer} \\
\midrule
1 & & 67.6 & 68.7 & 68.7 & & 27.3 & 26.3 \\
2 & & 67.9 & 65.7 & 35.8 & & 27.5 & 17.2 \\
3 & & 68.2 & 61.5 & 32.1 & & 25.1 & 12.7 \\
4 & & 66.8 & 50.8 & 27.7 & & 21.3 & 7.9 \\
5 & & 66.3 & 46.7 & 22.8 & & 19.3 & 6.5 \\
\bottomrule
\end{tabular}
\caption{Performance comparison between \sysname, Multi-layer, and CALM-DEC.}
\label{tab:benchmark}
\vspace{-1.7em}
\end{table}

As can be observed from Table \ref{tab:benchmark}, {\sysname} exhibits a superior performance over other approaches. This is demonstrated by the notably less degradation in task performance across both datasets as the speedup factor increases. This showcases the robustness of our method against increasing speedup.

\section{Related Work}

\noindent{\bf Model compression:} There has been extensive research in model compression to develop techniques to improve the inference efficiency of large language models (LLMs). One of the most prominent lines of work leverage knowledge distillation (KD)~\cite{DBLP:journals/corr/HintonVD15} to train smaller student models with faster inference using representations from LLMs as teachers like hidden states and attention states~\cite{jiao2019tinybert,sun2020mobilebert}. Another line of work in model compression use quantization~\citep{DBLP:journals/corr/GongLYB14}, low-precision training and network pruning~\citep{HanMao16}, parameter sharing and factorization to reduce the memory footprint as well as network latency~\citep{ compression_survey, efficient_survey}. Notably most of the above research in model compression has focused on encoder-only models for natural language understanding tasks.

\noindent{\bf Early-exit:} In contrast to the above works that use static computation i.e. the same computation being invoked on every input, we focus on {\em adaptive compute} with variable computation for  different parts of the input. Existing adaptive computation techniques primarily rely on early-exit strategies~\cite{DBLP:conf/acl/Zhu20,DBLP:conf/nips/ZhouXGM0W20,DBLP:conf/acl/XinTLYL20,DBLP:conf/acl/LiuZWZDJ20,DBLP:conf/emnlp/LiLCRLZS21,DBLP:conf/nips/HouHSJCL20} where a token in the input learns to exit from different layers of the network. Similar to the works in KD, most of these techniques were developed for encoder-only models like BERT~\cite{DBLP:conf/naacl/DevlinCLT19} for natural language understanding (NLU) tasks. In contrast to NLU tasks that requires processing of the sequences as a whole, generation tasks are more complex given their autoregressive nature for token-by-token generation. A recent work, CALM~\cite{schuster2022confident} study token-level early exit strategies for generation tasks in terms of what confidence measure to use; how to connect sequence-level constraints to local per-token exit decisions; and how to attend back to missing hidden representations due to early exits in previous tokens. However, similar to all the prior early-exit works, CALM suffers from some major practical blockers related to batching (only supporting a batch size of 1) and KV caching which are widely used to speedup inference in practice. Further, the worst-case scenario for all such exit-strategies (e.g., exit point closer to the top layer for any token) can lead to using the full network resulting in unpredictable system load and inconsistent throughput. To address these challenges, we develop {\sysname} that supports non-trivial batching and KV caching for efficient inference, as well as guarantees a predictable computational load with no surprises.

\section{Limitations and Future Directions}

{\sysname} addresses pivotal issues like batching and Key-Value (KV) caching being inherently incompatible with existing token-level early exit strategies. However, the introduction of the decaying policy adds a new limitation. As the generation progresses and samples in the batch finish their computations, new samples can be included in the batch only if their current position matches the remaining elements' positions. Therefore, our method does not naturally support the `infinite loop' inference mode.

In preliminary experiments, a power law decay function did not yield improvements over the linear decay employed in this study. Notably, prior research indicate a power law distribution for token exit levels \cite{schuster2022confident}. Our Oracle exploratory experiments, depicted in Figure \ref{fig:oracle}, corroborate this observation. Investigating alternative decay functions presents an intriguing avenue for future work.

Another promising research direction involves examining the decaying policy's impact on the prompt. In accordance with previous studies, we have utilized the full network for the prompt. Additional speedup gains may be attainable by extending the policy to the prompt and implementing more aggressive decay functions, as mentioned earlier. This could pave the way for more efficient and versatile token-level early exit strategies.

\section{Conclusions}

{\sysname} bridges the gap between the theoretical benefits of token-level early exits and real-world application requirements. It adeptly addresses practical challenges like batch processing and key-value caching. Moreover, it consistently exhibits the capability to reduce computational requirements by identifying the saturation point of hidden states with a controlled computational budget. This not only enhances efficiency but also fosters a more accessible and sustainable AI ecosystem. To further amplify these improvements, future efforts can concentrate on enhancing dynamic batching and delving deeper into the behavior of decay functions.

\bibliography{custom}
\bibliographystyle{plainnat}

\end{document}